\documentclass[letterpaper, 10 pt, conference]{ieeeconf}

\IEEEoverridecommandlockouts
\usepackage{graphics} 
\usepackage{epsfig} 
\usepackage{subfigure}

\usepackage{mathptmx} 
\usepackage{times} 
\usepackage{amsmath} 
\usepackage{amssymb} 
\usepackage[ruled,linesnumbered]{algorithm2e}
\usepackage{multirow}

\newtheorem{myrem}{Remark}
\newtheorem{theorem}{Theorem}
\newtheorem{assumption}{Assumption}
\usepackage{cite}
\usepackage[colorlinks=true,linkcolor=black,urlcolor=blue,citecolor=black]{hyperref}
\usepackage{bm} 
\usepackage{algorithmic}

\usepackage{color}

\begin{document}

\title{Trajectory Splitting: A Distributed Formulation for Collision Avoiding Trajectory Optimization    
}

\author{Changhao Wang$^1$, Jeffrey Bingham$^2$, and Masayoshi Tomizuka$^1$
\thanks{$^1$Dept. of Mechanical Engineering, University of California, Berkeley, CA, USA. 
{\tt\small \{changhaowang, tomizuka\} @berkeley.edu}}
\thanks{$^2$ X, the Moonshot Factory, Mountain View, CA, USA.
{\tt\small {jeffbingham@x.team}}
}}

\maketitle

\begin{abstract}

Efficient trajectory optimization is essential for avoiding collisions in unstructured environments, but it remains challenging to have both speed and quality in the solutions.
One reason is that second-order optimality requires calculating Hessian matrices that can grow with $O(N^2)$ with the number of waypoints. Decreasing the waypoints can quadratically decrease computation time. Unfortunately, fewer waypoints result in lower quality trajectories that may not avoid the collision. 
To have both, dense waypoints and reduced computation time, we took inspiration from recent studies on consensus optimization and propose a distributed formulation of collocated trajectory optimization. It breaks a long trajectory into several segments, where each segment becomes a subproblem of a few waypoints. 
These subproblems are solved classically, but in parallel, and the solutions are fused into a single trajectory with a consensus constraint that enforces continuity of the segments through a consensus update. With this scheme, the quadratic complexity is distributed to each segment and enables solving for higher-quality trajectories with denser waypoints. Furthermore, the proposed formulation is amenable to using any existing trajectory optimizer for solving the subproblems. We compare the performance of our implementation of trajectory splitting against leading motion planning algorithms and demonstrate the improved computational efficiency of our method.

\end{abstract}

\section{Introduction}

Finding optimal, collision-free trajectories is important for robots to interact with people and the environment. Sampling and optimization are two of the most powerful ways to achieve the goal. Optimization allows defining the problem in terms of constraints and finding solutions that optimize performance. Currently, collocation-based optimization methods, such as TrajOpt~\cite{schulman2013finding}, solve a non-linear program (NLP) using non-linear optimization algorithms like sequential quadratic programming (SQP)~\cite{boggs1995sequential}. The structure of the NLP leads to at least $O(N^2)$ time complexity concerning the number of waypoints to satisfy second-order optimality criteria. Therefore, increasing the density of waypoints quadratically increases the computation time. In comparison, sampling-based planners, such as rapidly-exploring random trees (RRT)~\cite{lavalle2006planning}, are able to be parallelized to achieve a higher computation efficiency. 

Our goal is to combine the expressiveness and quality of optimization-based techniques with the computational advantages of parallelization in sampling-based methods. 
We propose a method that separates the trajectory optimization problem into a set of subproblems that can be solved in a distributed manner.

Prevailing methods for parallel or distributed optimization mainly focus on multi-agent planning~\cite{rey2018fully, li2020energy} with limited research applying the techniques to single-agent planning. Brendan. et al~\cite{o2013splitting} pioneered in this field by applying the operator splitting method to convex optimal control problems. They proposed splitting an optimal control problem into subproblems by constraint. This idea is further studied in~\cite{sindhwani2017sequential} and applied to more challenging control scenarios.
If it were applied to a collision-avoiding trajectory problem for a robot, two subproblems would be created. One subproblem would satisfy only the robot dynamics and the other would only avoid the collision. These problems would then be fused using an iterative consensus update to find the optimal trajectory. However, even with this splitting scheme, the number of variables for each subproblem remains the same, and it still suffers from the $O(N^2)$ time complexity. Furthermore, splitting by constraint may create subproblems with different complexity. Some subproblems may dominate the computation time, and others should wait until the most complicated subproblem is finished in order to begin the next consensus iteration. 

To deal with those problems, we propose to split the problem by the path variables, creating subproblems for segments of the trajectory. In this way, the problem complexity can be equally distributed. These segments are then fused together using a consensus update scheme similar to the method described above. This proposed method of “trajectory splitting” exploits the observation that complexity decreases as path-length decreases and in this way offers a more efficient approach of solving trajectory optimization for certain classes of problems. Moreover, the proposed splitting scheme is amenable to be incorporated with any existing trajectory optimizers to solve the subproblem.
To the authors' knowledge, this is the first approach to parallelizing optimization-based trajectory planning by splitting the path variables in order to equally distribute the problem complexity to each trajectory segment.
The contributions of the proposed method are listed as follows:
\begin{itemize}
    \item A novel formulation of trajectory optimization problems via splitting the trajectory into segments that can equally distribute the problem complexity.
    \item A distributed optimization algorithm to solve the proposed formulation for better computational efficiency.
    \item An implementation of the trajectory splitting algorithm with a state-of-the-art collision checker and optimization solver.
    \item Comprehensive comparisons of the proposed trajectory splitting algorithm against existing methods using both simulation and real-world experiments.
\end{itemize}

\section{Related Works}

Motion planning algorithms are nominally classified into two broad categories, sampling, and optimization. Sampling methods are well suited for problems where any feasible solution is acceptable and gradient information is difficult or expensive to compute. Probabilistic roadmap (PRM)~\cite{kavraki1996probabilistic} solves the planning problem by constructing a complete map of the workspace and then using a search algorithm to find a feasible path from the map. One major problem of PRM is the computation time. It is time-consuming to build a complete roadmap, especially in high-dimensional space. RRT~\cite{lavalle2006planning} deals with this problem by incrementally building a graph and checking feasible paths at the same time. In practice, RRT and its variants~\cite{kuffner2000rrt, lavalle2006planning} are still some of the most powerful ways to deal with the planning problem. They are efficient and can easily be parallelized~\cite{lavalle2006planning} for even better performance. However, since sampling-based planners are stochastic, they may find different solutions for the same problem, and the solution quality and the computation time may also have a large variance. This problem has been reported in several papers~\cite{zucker2013chomp, meijer2017performance}. It is still an active research area to improve the robustness of sampling planners.

Though sampling planners are effective to find feasible paths, it is often preferable to obtain `optimal' paths that satisfy an objective.
RRT*~\cite{lavalle2006planning}, an optimal variant of traditional RRT, is able to obtain optimal paths with the help of an additional `rewire' operation. Exploiting the probabilistic completeness property provides a theoretical guarantee of globally optimal solutions. However, in practice, the planner is only given finite time to find a path and thus, RRT* performance declines sharply as the dimensionality of the scenario increases~\cite{meijer2017performance}.

Optimization, on the other hand, provides a way to go beyond finding a feasible solution and offers a means to seek `better' solutions based on an objective. State-of-the-art trajectory optimization algorithms start with an infeasible solution and evolve a trajectory to minimize a defined cost and satisfy all constraints. CHOMP~\cite{zucker2013chomp} pioneered this approach for collision avoidance planning problems in robotics, proposing a covariant gradient descent update rule to optimize the trajectory. To deal with non-differentiable constraints, STOMP~\cite{kalakrishnan2011stomp} proposed a stochastic update rule. TrajOpt~\cite{schulman2013finding} introduced an SQP formulation to solve the planning problems. Recently, there has been considerable progress in this field~\cite{mukadam2018continuous, liu2018convex}, and these efforts suggest new approaches to optimization-based planning has the potential to be both computationally efficient and retain high-quality solutions.

\section{Mathematical Background}
In this section, we introduce the preliminaries of the proposed method. The classic trajectory optimization formulation by collocation is introduced in Section~\ref{subsection::NLP_formulation}. The mathematical background of consensus optimization and alternating direction method of multipliers (ADMM) is illustrated in Section~\ref{subsection::admm}.

\subsection{Trajectory Optimization Formulation}
\label{subsection::NLP_formulation}

A common discrete form of trajectory planning can be formulated as an optimization problem with the following form:
\begin{equation}
\begin{aligned}
\min_{\tau} \quad & c(\tau)\\
\textrm{s.t.} \quad & x_{i+1} = f(x_i)\\
& g(x_i) \leq 0\\
& i=1,\cdots,N-1
 \label{eqn_trajectory_optimization_formulation}
\end{aligned}
\end{equation}
where $\tau = \{x_1,\cdots,x_N\}$ denotes a discretized robot trajectory, and $x_i$ denotes the robot state (position, velocity, and acceleration). $c(\cdot)$ is a human designed cost function, $f(\cdot)$ may include the robot kinematics or dynamics constraints, and $g(\cdot)$ is the robot state constraint. In this paper, we assume the cost function and constraints are separable (or block-separable) $c(\tau) = \sum_{i=1}^{N} c_i(x_i)$.

\subsection{ADMM and Consensus Optimization}
\label{subsection::admm}

Consensus optimization~\cite{boyd2011distributed} considers the problem with separable objectives:
\begin{equation}
\begin{aligned}
\min_{x_1,\cdots,x_N,z \in \mathbb{R}^N} \quad & \sum_{i=1}^N c_i(x_i)\\
\textrm{s.t.} \quad
  & x_i =z \quad i = 1, \cdots, N\\
 \label{eqn_consensus_formulation}
\end{aligned}
\end{equation}
where $x_i$ is called a local variable, and $z$ is a global value that each local variable tries to achieve.

ADMM~\cite{boyd2011distributed}, an augmented Lagrangian method~\cite{bertsekas1997nonlinear}, is able to parallelize the consensus problem and solve it efficiently. The augmented Lagrangian of (\ref{eqn_consensus_formulation}) is shown in (\ref{eqn::augmented_lagrangian}), where $y_i$ denotes the Lagrange multiplier of the corresponding consensus constraint.
Similar to penalty methods, the augmented Lagrangian adds an additional constraint term to the original Lagrangian in order to penalize the constraint violation, and $\rho$ is a weight that controls the constraint violation.

\begin{equation}
\begin{aligned}
 L &= \sum_{i=1}^N [c_i(x_i) + y_i^T(x_i-z) + (\rho/2) \|x_i-z\|^2\\
\end{aligned}
\label{eqn::augmented_lagrangian}
\end{equation}

\begin{figure}
    \centering
    \includegraphics[scale=0.32]{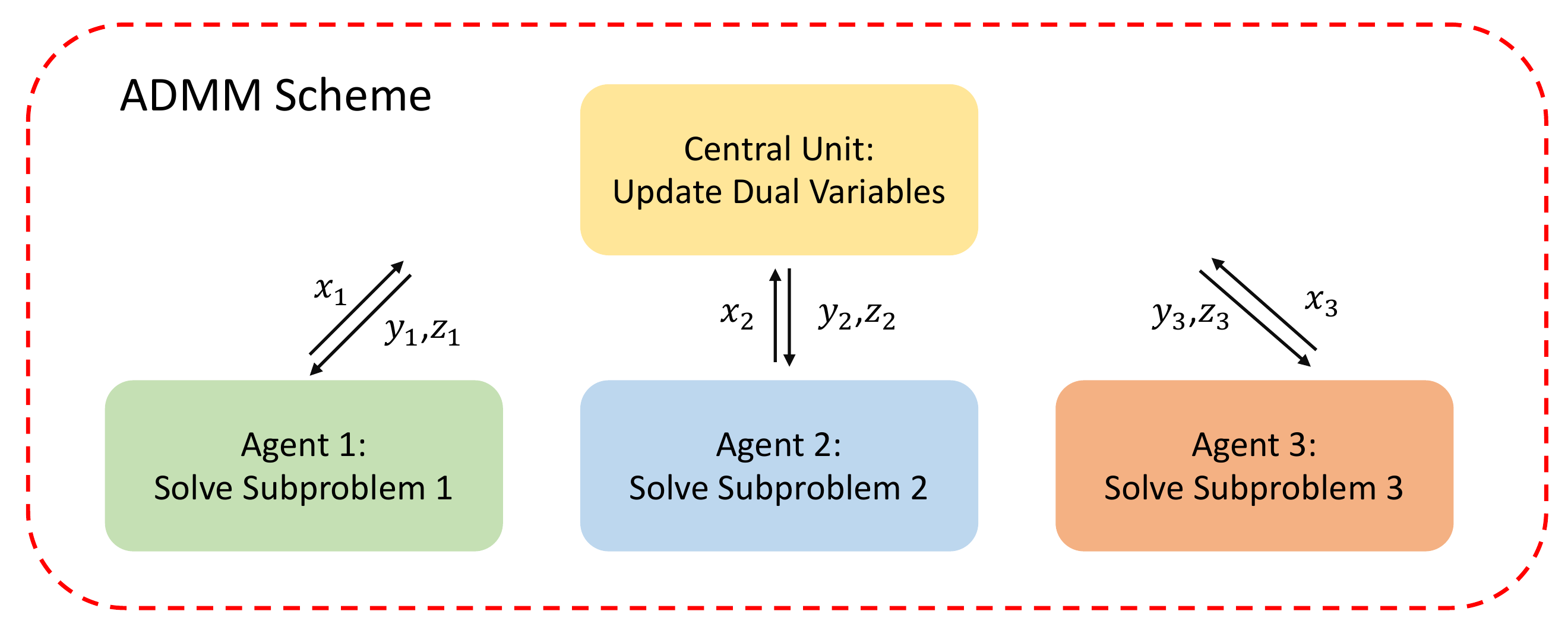}
    \caption{An illustration of the ADMM update rule to solve consensus problems. ADMM solves the problem in a distributed manner, where it creates $N$ separate agents to solve each subproblem. A central unit then collects the results obtained from each agent and updates the dual variable until a stopping criterion is met.}
    \label{fig:ADMM}
\end{figure}

Based on the augmented Lagrangian in (\ref{eqn::augmented_lagrangian}), ADMM can solve the original problem in a distributed way as shown in Fig.~\ref{fig:ADMM}. First, ADMM initializes
$N$ separate agents (solver) in order to update each local variable $x_i$. Then, a central unit collects the solution from the agent and updates the dual variable accordingly $y_i$. The update rule is then given by:
\begin{equation}
\begin{aligned}
     x_i^{k+1} &= \textrm{arg}\min_{x_i} \quad [c_i(x_i) + y_i^{kT}(x_i - z^k) + (\rho/2)||x_i - z^k||^2] \\
     y_i^{k+1} & = y_i^{k} + \rho (x_i^{k+1} + z^{k+1}) \\
     z^{k+1} & = \frac{1}{N} \sum_{i=1}^N x_i^{k+1}
\end{aligned}
\label{eqn::consensus_update_rule}
\end{equation}

The convergence and optimality condition of ADMM is illustrated in Theorem 1. For the convex consensus optimization problem, a globally optimal solution is guaranteed. 

The ADMM algorithm has also been extensively applied to non-convex, coupling optimization problems~\cite{wang2019global, liu2019linearized}. Recently, convergence criteria for general non-convex problems have been studied and we refer readers to~\cite{wang2019global} for further details. We will discuss convergence for our implementation and similar problems in Section~\ref{subsection::convergence}. In summary, ADMM algorithms show amazing practical success in solving both convex and general nonlinear optimization problems even while the theoretical proof of convergence is forthcoming.

\begin{assumption}
$c_i(x) ,i= 1,\cdots,N$ are closed, proper, and convex.
\end{assumption}
\begin{assumption}
The augmented Lagrangian $L_0$ contains a saddle point
\end{assumption}

\begin{theorem}
Under Assumptions 1 and 2, the ADMM iteration satisfies the following~\cite{boyd2011distributed}:
\begin{itemize}
    \item Residual convergence: $x_i - z \rightarrow 0$ as $k \rightarrow \infty$
    \item Objective convergence: $\sum_{i=1}^N c_i(x_i) \rightarrow f^*$ as $k \rightarrow \infty$
    \item Dual convergence: $y_i^k \rightarrow y_i^*, i=1,\cdots,N$ as $k\rightarrow \infty$
\end{itemize}
\end{theorem}

It is worth noticing that (\ref{eqn_consensus_formulation}) is a simplified form of the consensus problem. For the general form~\cite{boyd2011distributed},
each local variable, or even each element, can have its own consensus constraints. The above update rule can be easily transformed to the general case by replacing the consensus constraint, and the optimality and convergence analysis still hold for the general formulation.

\section{Trajectory Splitting Algorithm}

Trajectory optimization aims at finding a smooth and collision-free trajectory between two predefined states. Similar to previous methods \cite{schulman2013finding}, we formulate the motion planning problem as an NLP as introduced in Section~\ref{subsection::NLP_formulation}. 
While in contrast with other methods, which directly apply optimization algorithms (such as, covariant gradient descent or SQP) to solve the entire problem, we propose splitting the trajectory into several segments and solving them in a distributed manner for better computational efficiency. The splitting formulation as a consensus optimization will be introduced in Section~\ref{subsection::three_waypoints_example}, \ref{subsection::general_splitting_formulation}. The collision avoidance constraints will be formalized in Section~\ref{subsection::collision_constraints}. The optimization update rule will be explained in Section~\ref{subsection::optimization}. In the end, we provide the stopping criterion and convergence analysis of the proposed algorithm.

\subsection{Intuition for Trajectory Splitting: A Three Waypoint Example}
\label{subsection::three_waypoints_example}
\begin{figure}
    \centering
    \includegraphics[scale=0.27]{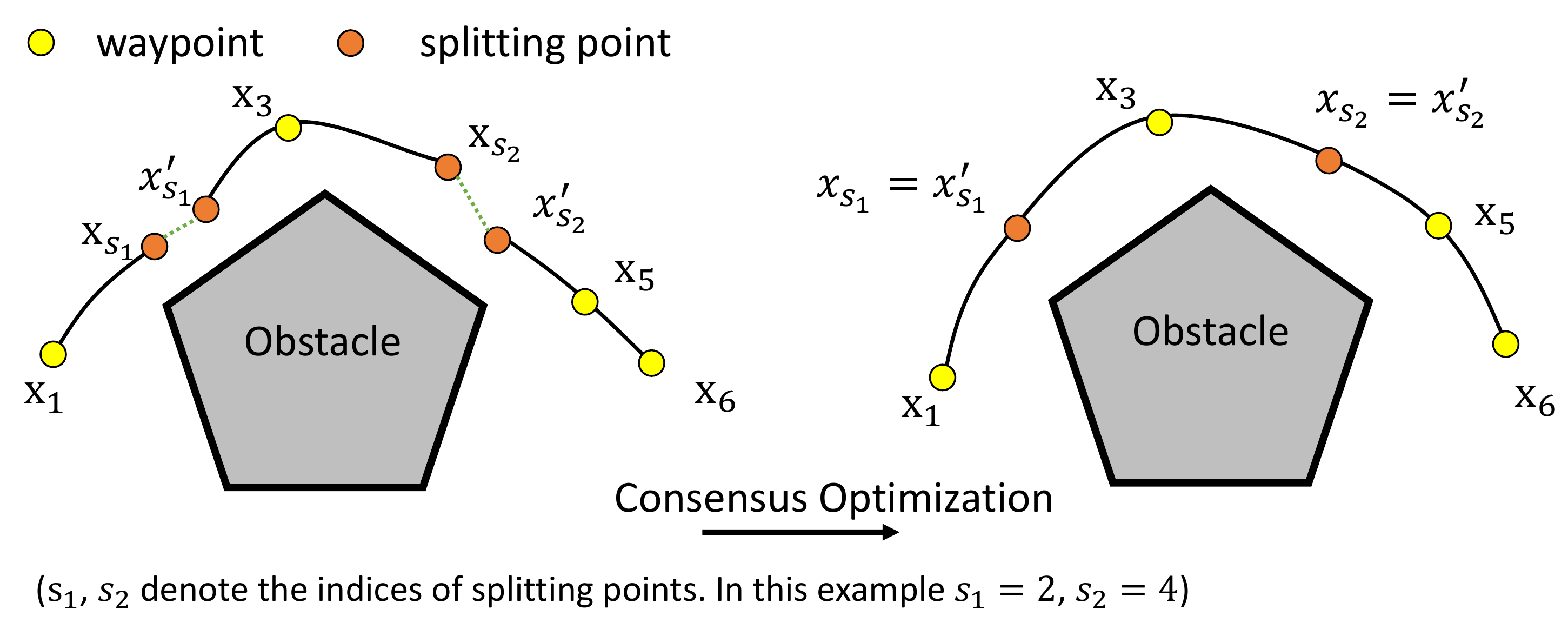}
    \caption{Illustration of trajectory splitting algorithm. The waypoints are denoted by the yellow dots, and the orange dots are the splitting points $x_{s_i}$. The trajectory splitting algorithm is able to find feasible trajectory pieces in parallel and then connect them together.}
    \label{fig:splitting_illustration}
\end{figure}

Let us begin with an example with three waypoints $x_1, x_2,x_3$. Consider we want to split the trajectory at $x_2$ creating a leading trajectory of $\tau_1 = \{x_1,x_2\}$, and a trailing trajectory of $\tau_2 = \{ x_2^\prime, x_3\}$. Here $x_2^\prime$ is a slack variable that is defined to be equal to $x_2$. According to the consensus formulation in (\ref{eqn_consensus_formulation}), in order to achieve the consensus between $x_2$ and $x_2^\prime$, a global variable $z$ is introduced to enforce this constraint. Therefore, we can rewrite the three waypoint trajectory optimization problem in (\ref{eqn_trajectory_optimization_formulation}) as follows:
\begin{equation}
\begin{aligned}
\min_{x_1,x_2,x_2^\prime,x_2,z} \quad & c_1(x_1) + \frac{1}{2}c_2(x_2) + \frac{1}{2}c_2(x_2^\prime)+ c_3(x_3)\\
\textrm{s.t.} \quad & x_2 = z,\quad  x_2^\prime = z\\
  & x_{2} = f(x_1), \quad x_{3} = f(x_2^\prime)\\
  & g(x_i) \leq 0, \quad i = 1,2,3\\
  & g(x_2^\prime) \leq 0\\
 \label{eqn_split_example}
\end{aligned}
\end{equation}

Notice that the problem is separable between these two trajectory pieces, except for the first two consensus constraints. 
According to the consensus optimization update rule in (\ref{eqn::consensus_update_rule}), each trajectory segment $\tau_1$, and $\tau_2$ can be updated by solving the following optimization where the dual variable update follows the same manner as in (\ref{eqn::consensus_update_rule}).

\begin{equation}
\begin{aligned}
\min_{x_1,x_2} \quad & c_1(x_1) + \frac{1}{2}c_2(x_2) + y_1^T(x_2 - z)+ (\rho/2) \|x_2 - z\|^2\\
\textrm{s.t.} \quad & x_{2} = f(x_1)\\
  & g(x_1) \leq 0, \quad i = 1,2\\
 \label{eqn_split_example_sub_1}
\end{aligned}
\end{equation}

\begin{equation}
\begin{aligned}
\min_{x_2^\prime,x_3} \quad & \frac{1}{2}c_2(x_2^\prime) + c_3(x_3)+ y_2^T(x_2^\prime - z) + (\rho/2) \|x_2^\prime - z\|^2\\
\textrm{s.t.} \quad & x_{3} = f(x_2^\prime)\\
  & g(x_2^\prime) \leq 0\\
  & g(x_3) \leq 0\\
 \label{eqn_split_example_sub_2}
\end{aligned}
\end{equation}

\begin{myrem}
Constraints can be incorporated into the objective function via indicator functions. Therefore, it does not affect the update rule for the ADMM formulation.
\end{myrem}

\subsection{General Trajectory Splitting Formulation}
\label{subsection::general_splitting_formulation}

\begin{algorithm}
\caption{Trajectory Splitting}
\begin{algorithmic}[1] 
\label{alg_TrajSplit}
\REQUIRE Trajectory: $\tau_i$, Dual variables: $y_{i,1}$, $y_{i,2}$
\WHILE{splitting tolerance (\ref{eqn::splitting_tolerance}) not satisfied}
    \FOR {$i=1,\cdots,M+1$}
        \STATE $\tau_i \leftarrow$ Solve Eq. \ref{eqn_split_subproblem}
    \ENDFOR
    \FOR {$i=1,\cdots,M$}
        \STATE $z_i = \frac{1}{2}(x_{s_i} + x_{s_i}^\prime)$
        \STATE $y_{i,1} = y_{i,1} + \rho(x_{s_i} - z_i)$
        \STATE $y_{i,2} = y_{i,2} + \rho(x_{s_i}^\prime - z_i)$
    \ENDFOR
\ENDWHILE
\RETURN $\tau_i\quad i=1,\cdots,M+1$
\end{algorithmic}
\end{algorithm}

Consider we split the trajectory into $M+1$ segments, and the position that splitting happens is denoted by $x_{s_i}$, where $s_i$ is the index of the $i~$th splitting point on the original trajectory as shown in Fig. \ref{fig:splitting_illustration}. Similar to the previous example, we introduce the slack variable $x_{s_i}^\prime$ as a copy of the splitting point. In order to enforce that the trajectories are connected with each other, a global variable $z_i$ is used for this constraint:
\begin{equation}
\begin{aligned}
x_{s_i} &= z_i \\
x_{s_i}^\prime &= z_i, \quad i=1,\cdots,M
\end{aligned}
\label{eqn::spliting_constraint}
\end{equation}
and we use $y_{i,1}$ and $y_{i,2}$ to denote the Lagrangian multipliers of the consensus constraints above.  
Thus, we can formulate the augmented Lagrangian $L_i$ for each trajectory piece as shown in (\ref{eqn::Lagrangian}).
\begin{figure*}
\normalsize
\begin{equation}
\begin{aligned}
    L_1 &= \sum_{j=1}^{s_1-1} c_{j}(x_{j}) + \frac{1}{2}c_{s_1}(x_{s_1}) + y_{1,1}^T(x_{s_1} -z_1) + (\rho/2) \|x_{s_1} - z_1\|^2 \\
    & \cdots \\
    L_i & = \frac{1}{2}c_{s_{i-1}}(x_{s_{i-1}}^\prime) + \sum_{j=s_i+1}^{s_{i+1} - 1} c_{j}(x_{j}) + \frac{1}{2}c_{s_i}(x_{s_i})
   + y_{i-1,2}^T(x_{s_{i-1}}^\prime - z_{i-1}) + (\rho/2) \|x_{s_{i-1}} - z_{i-1}\|^2
   + y_{i,1}^T(x_{s_i} - z_{i}) + (\rho/2)\|x_{s_i}- z_i\|^2\\
   & \cdots \\
L_{M} & = \sum_{j=s_{M}+1}^{N} c_{j}(x_{j}) + \frac{1}{2}c_{s_M}(x_{s_M}^\prime) + y_{M,1}^T(x_{s_M}^\prime -z_M) + (\rho/2) \|x_{s_M}^\prime - z_M\|^2
\end{aligned}
\label{eqn::Lagrangian}
\end{equation}
\hrulefill
\end{figure*}
Then the primal update rule for each trajectory segment is given in (\ref{eqn_split_subproblem}).
\begin{equation}
\begin{aligned}
\tau_i^{k+1} = \min_{\tau_i} \quad & L_i\\
\textrm{s.t.} \quad & x_{j+1} = f(x_{j})\\
  & g(x_{j}) \leq 0\\
  & \forall x_{j} \in \tau_{i}\\
\label{eqn_split_subproblem}
\end{aligned}
\end{equation}

The overall algorithm is illustrated in Alg.~\ref{alg_TrajSplit}. In practice, the trajectory $\tau$ can be initialized as a straight line that goes from the initial point to the target point, or it can be given as a feasible solution from sampling-based planners. The Lagrange multipliers $y_{i,1}$, $y_{i,2}$ can be initialized as zero vectors. The number of splitting $M$ could be any integer that is smaller than the number of waypoints, and the waypoints are then uniformly separated into each subproblem.

\subsection{Collision Avoidance Constraints}
\label{subsection::collision_constraints}

\begin{figure}
    \centering
    \includegraphics[scale=0.28]{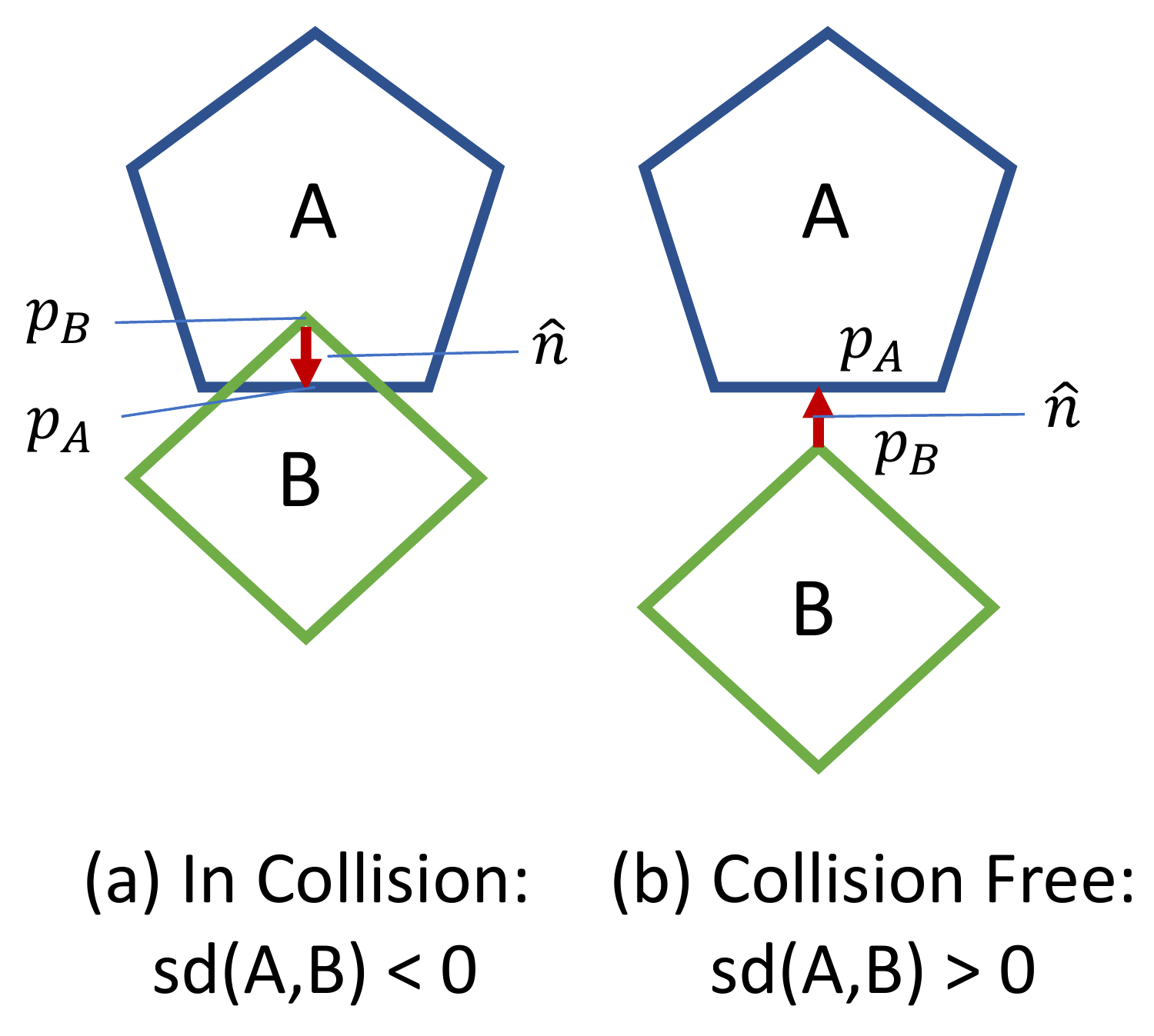}
    \caption{Illustration of signed distance function. The signed distance value is positive if objects are collision free, and the value is negative if they are in collision.}
    \label{fig:signed_distance}
\end{figure}

We consider collision-avoidance constraint $g(x_i) \leq 0$ as a function of signed distance~\cite{schulman2013finding}.
As shown in Fig.~\ref{fig:signed_distance}, the signed distance function denotes the minimum distance between two objects, where the sign is determined by whether the surfaces of the objects interpenetrate. We mathematically define the signed distance function between two objects $A$ and $B$ by:
\begin{equation}
    \text{sd}(A,B) = \text{dist}(A,B) - \text{penetration}(A,B)
\end{equation}
where the `dist' function is defined as the minimum translation distance to just cause contact between the surfaces of a pair of objects:
\begin{equation}
    \text{dist} = \inf \{\|T\|: \exists p_A \in A, p_A + T \in B\}
\end{equation}
and similarly, the `penetration' function denotes the minimum translation that moves the two objects out of contact:
\begin{equation}
    \text{penetration} = \inf \{\|T\|: \forall p_A \in A, p_A + T \notin B\}
\end{equation}

In our implementation, we approximated the signed distance using methods from the Flexible Collision Library (FCL)~\cite{pan2012fcl}. In turn, this signed distance method was used to formulate a  collision-avoidance constraint where the signed distance between the robot and each obstacle had a strictly positive value along the entire trajectory.
An additional approximation was adopted by utilizing the Jacobian for the contact point from the signed distance, as introduced in~\cite{schulman2013finding}. With this technique, assume the object $A$ is a robot link and its position is determined by the robot state $x$. Then, the signed distance is defined by contact points denoted by $p_A$ for the robot link and $p_B$ for an obstacle. A static assumption is made to arrive at the approximation given in (\ref{eqn::signed_distance_approx}) by assuming the contact point $p_A$ is not a function of $x$: 

\begin{equation}
    sd_{AB}(x) \approx \hat{n}(F_A^w(x)p_A - F_B^wp_B)
\label{eqn::signed_distance_approx}
\end{equation}
where $F_A^w$ and $F_B^w$ are the homogeneous transformation from $A$, and $B$ frames to the world coordinate, and $\hat{n}$ is the direction that points from $p_B$ to $p_A$.

In addition, the gradient of this constraint can be computed as a Jacobian of the signed distance $\text{sd}_{AB}$ with the following approximation:
\begin{equation}
    \nabla \text{sd}_{AB}(x) \approx \hat{n}^T J_{p_A}(x)
\end{equation}
where $J_{p_A}$ denotes the robot translation Jacobian at $p_A$.

\subsection{Optimization}
\label{subsection::optimization}

The key to this algorithm is the separation of the trajectory optimization problem into several subproblems that can be solved independently. 
For each subproblem in (\ref{eqn_split_subproblem}), it is expressed in a standard form for trajectory optimization; therefore, existing optimization algorithms, such as gradient descent, SQP~\cite{boggs1995sequential}, CHOMP~\cite{zucker2013chomp}, and TrajOpt~\cite{schulman2013finding} can be directly applied. 

In our implementation, we use the optimization library IPOPT~\cite{wachter2006implementation} as the basis of our subproblem solver. IPOPT implements a primal-dual interior-point filter line search algorithm to solve the general nonlinear optimization problems, and the convergence of this algorithm is proved in~\cite{wachter2005line}.

\begin{figure*}[!ht]
\centerline{\includegraphics[scale = 0.44]{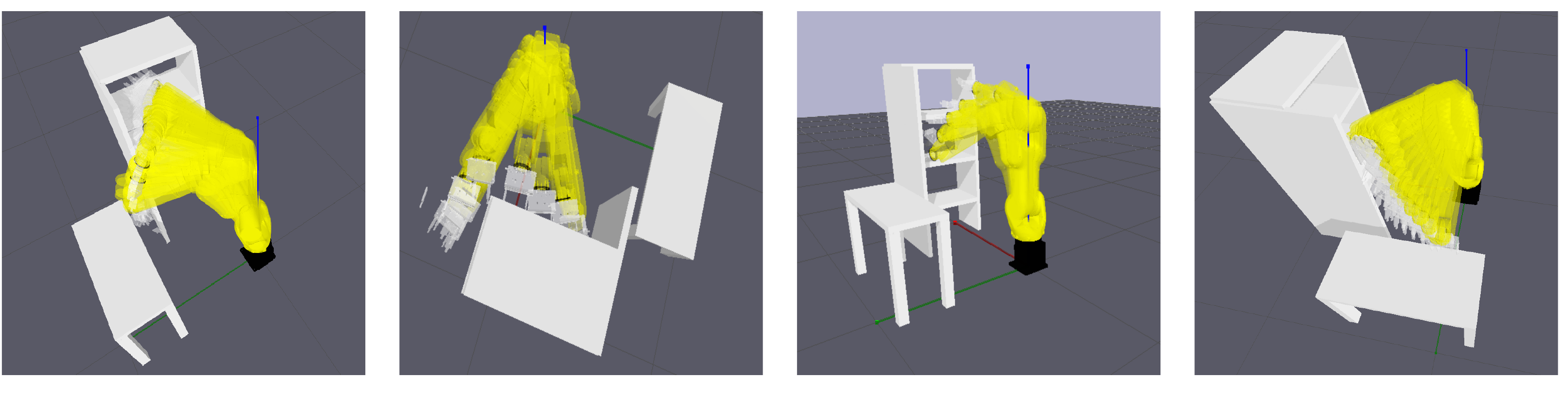}}
\caption{Simulation benchmark results showing four different planning problems and the results obtained by the trajectory splitting algorithm in the bookcase scenario. The algorithm is able to find the local optimal solution for each trajectory segment efficiently and connect them together in a smooth trajectory.
}
\label{fig_snapshots}
\end{figure*}

\subsection{Stopping Criterion and Convergence Analysis}
\label{subsection::convergence}
Similar to~\cite{zhou2020accelerated}, we choose to use the primal residuals as the stopping criterion:
\begin{equation}
    \|r^k\|_2 \leq \epsilon
\label{eqn::splitting_tolerance}
\end{equation}
where $\epsilon$ is a predefined positive scalar, and splitting tolerance $r^k$ is the average of constraint violations:
\begin{equation}
    \|r^k\|_2 = \frac{1}{M} (\sum_{i=1}^M \|q_{s_i}^k - {q_{s_i}^\prime}^k\|^2)^{\frac{1}{2}}
\end{equation}

The convergence of the proposed trajectory splitting algorithm can be divided into three cases:
\begin{itemize}
    \item \textbf{Convex}: According to Theorem 1, if the objective functions and constraints are convex (the equality constraint should be linear), the proposed trajectory splitting algorithm is guaranteed to converge to a global optimal solution. An example of this type of problem is the Linear-Quadratic Regulators (LQR). For the LQR problem, the trajectory splitting algorithm is guaranteed to find a global optimal solution.
    
    \item \textbf{Restricted prox-regular\cite{wang2019global}}: Recent studies provide the convergence of ADMM on non-convex, non-smooth problems. \cite{wang2019global} proves the condition that if the objectives and constraints satisfy the restricted prox-regular condition, then the ADMM algorithm will converge to a stationary point. In our scenario, if the dynamics constraint $f(\cdot)$ is linear, and the nonlinear collision constraint $g(\cdot)$ is restricted prox-regular, then the solution $(\tau, y, z)$ obtained from Alg.~\ref{alg_TrajSplit} will converge to a stationary point $(\tau^*,y^*,z^*)$ for the augmented Lagrangian in (\ref{eqn::Lagrangian}).
    According to \cite{wang2019global}, restricted prox-regular is a weaker condition than prox-regular and semi-convex\cite{colesanti2000hessian}. As studied in~\cite{liu2018convex}, the signed distance function between two convex objects is semi-convex; therefore, a large set of trajectory optimization problems fall into this category. This includes mobile robot planning with convex obstacles and linear kinematics (as shown in Section~\ref{subsection:2d_simulation}).
    
    \item \textbf{General non-convex}: For the general non-convex case, the convergence of ADMM is still an active field of optimization research. The multi-joint robotic motion planning problem falls into this category. Though the signed distance function of convex obstacles satisfies the semi-convex condition, the nonlinear robot forward kinematics violates these conditions and makes the planning problem extremely challenging. This is a common problem encountered by all the existing trajectory optimization algorithms. Though there is no theoretical proof of convergence yet, our practical observation from simulation and experiments show that the trajectory splitting algorithm converges reasonably for this challenging scenario (as shown in Section~\ref{subsectin::multi_simulation}, and Section~\ref{subsection::Experiments}).
\end{itemize}

\section{Simulation and Experiments}

\begin{figure}
    \centering
    \includegraphics[scale=0.38]{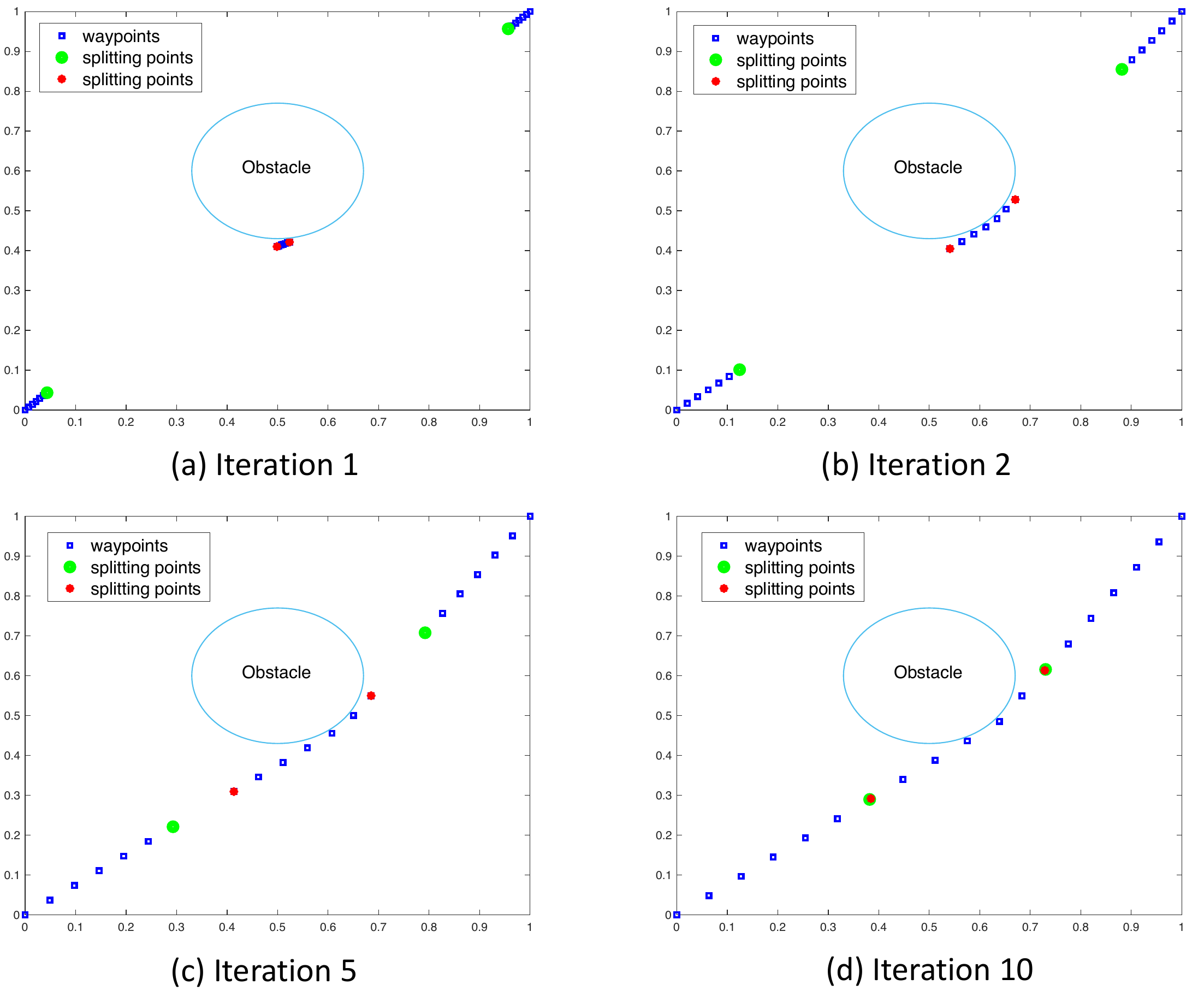}
    \caption{A 2D example of trajectory splitting. The planning problem is split into three segments and solved in a distributed manner. The trajectory is initialized using linear interpolation. As the optimization begins, each segment is very short and there is a large splitting error since the individual problems have an objective that minimizes segment distance. As the optimization progresses, the consensus update reduces the splitting error and a continuous, smooth trajectory is obtained.}
    \label{fig:2d_example}
\end{figure}

\begin{table*}
\centering
\caption{Simulation Benchmark of 6-DOF Robot Planning}
\label{table1}
\begin{tabular}{l l l l l l l l}
\hline
Algorithm  & RRT & LBKPIECE & RRT* & CHOMP & IPOPT & \textbf{TrajSplit\_5} & \textbf{TrajSplit\_3}\\ \hline
Average Time (s) & $0.254$ &$0.218$ & $5.001$ & $0.267$ & $0.489$ & $0.381$ & $0.178$ \\ \hline
Path Length (rad) & $11.24$ & $10.97$ & $9.03$ & $7.15$ & $7.23$ & $7.82$ & $7.43$ \\ \hline
Success Rate  & $25/25$ & $25/25$ & $25/25$ & $21/25$ & $23/25$ & $21/25$ & $23/25$ \\ \hline
\end{tabular}
\end{table*}

\subsection{Robot Model}
The proposed trajectory splitting algorithm is tested in two scenarios. The first scenario is a simple 2D case with a sphere obstacle (as shown in Fig.~\ref{fig:2d_example}), and the second is on a 6-DoF FANUC LRMATE 200iD robot (as shown in Fig.~\ref{fig_snapshots}). 

For the 2D example, the robot state $x$ is defined as the Cartesian position, and velocity in 2D: $x = [p_x, p_y, \dot p_x, \dot p_y]^T$, where $p_x$, and $p_y$ denote the position in $x$ and $y$ axes. 

For the multi-jointed robot case, the robot state $x$ is selected to include the robot joint angle $\theta \in \mathbb{R}^6$, joint velocity $\dot \theta \in \mathbb{R}^6$, and joint acceleration $\ddot \theta \in \mathbb{R}^6$. For optimization, we use linear double-integrator dynamics as the equality constraint:
\begin{equation}
    \begin{bmatrix}
    \theta_{i+1}\\
    \dot \theta_{i+1}
    \end{bmatrix}\ = 
    \begin{bmatrix}
    \bm{I} & T \bm{I}\\
    \bm{0} & \bm{I}
    \end{bmatrix}
    \begin{bmatrix}
    \theta_i\\
    \dot \theta_i
    \end{bmatrix}
    +
    \begin{bmatrix}
    \bm{0}\\
    T\bm{I}
    \end{bmatrix} \ddot \theta_i
\label{eqn::linear_double_integrator_dyanmics}
\end{equation}
where $\bm{0} \in \mathbb{R}^{6\times6}$, and $\bm{I} \in \mathbb{R}^{6\times 6}$ are the zero and identity matrices, and $T$ is a positive scalar that denotes the robot travel time in-between each consecutive waypoint pair.
The objective function is selected to minimize the summation of joint velocities in (\ref{eqn::objective_function}) for the minimum path length trajectory. 

\begin{equation}
    c(x) = \sum_{i=1}^N \|\dot \theta_i\|^2
    \label{eqn::objective_function}
\end{equation}

\subsection{Simulation in 2D}
\label{subsection:2d_simulation}

We first evaluate the effectiveness of the proposed trajectory splitting algorithm in a 2D scene with a sphere obstacle. The collision avoidance constraint is formulated as:
\begin{equation}
    ||\begin{bmatrix}
    p_x\\
    p_y
    \end{bmatrix}
    - \begin{bmatrix}
    {p_{o}}_x\\
    {p_{o}}_y
    \end{bmatrix}||^2 \geq r_{o}^2
\end{equation}
where $p_o$ and $r_o$ denote the obstacle center position and its radius. As we can see from the simulation results in Fig.~\ref{fig:2d_example}, the trajectory is split into three segments. The evolution of the solution initially produces short segments with a large splitting error that is refined through the consensus iteration to produce the final smooth trajectory.

\subsection{Simulation on Multi-jointed Robot}
\label{subsectin::multi_simulation}

\begin{figure}
    \centering
    \includegraphics[scale=0.42]{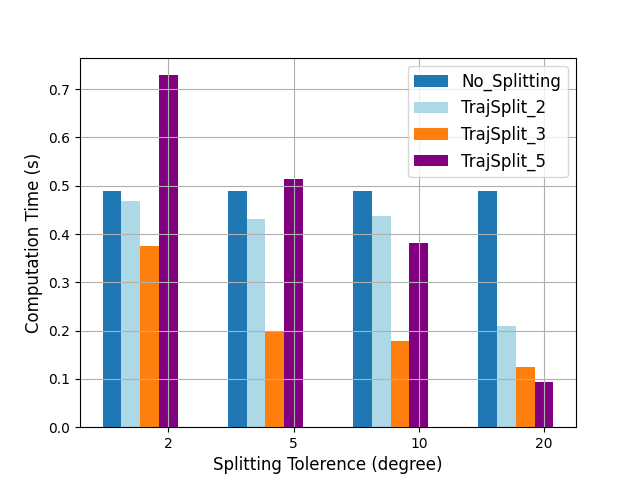}
    \caption{Parameter sweeps for the trajectory splitting algorithm, where Trajsplit\_2, Trajsplit\_3, Trajsplit\_5 denote using the splitting algorithm to separate a trajectory into two, three, and five pieces. Computation time decreases as splitting error tolerance in the stopping criterion is increased. In general, increasing the number of split segments does not guarantee a shorter computation time. For these problems, three segments provide a nice balance between the number of ADMM iterations and subproblem complexity.}
    \label{fig:splitting_error_time}
\end{figure}

\begin{figure}
    \centering
    \includegraphics[scale=0.42]{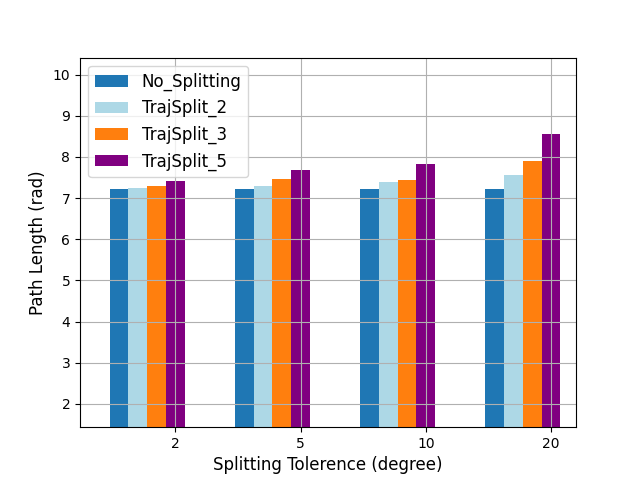}
    \caption{Parameter sweeps for the trajectory splitting algorithm, where Trajsplit\_2, Trajsplit\_3, Trajsplit\_5 denote using the splitting algorithm to separate a trajectory into two, three, and five pieces. Reducing the splitting error tolerance results in smoother trajectories at the cost of greater computation time. These problems show a good reasonable trade-off in quality and performance by setting the splitting error tolerance to $10$ degrees with two or three split segments.}
    \label{fig:splitting_error_length}
\end{figure}

We benchmarked the proposed trajectory splitting algorithm against existing state-of-the-art motion planners and optimization solvers, which include sampling-based methods RRT, RRT*, LBKPIECE in OMPL/Moveit, and an optimization-based method CHOMP in MoveIt using default parameters. We also implemented an optimization baseline using IPOPT with FCL as the collision checker. Our implementation of the trajectory splitting planner is based on the IPOPT baseline and uses the python multiprocess library to achieve parallelization. 

Fig. \ref{fig_snapshots} shows the 4 of the benchmark results obtained from the proposed trajectory splitting algorithm. We manually selected $5$ start poses and $5$ end poses to construct $25$ unique planning problems. The planning time limit is set to $5$ seconds.
Since MoveIt planners can only compute paths instead of trajectories, to make the comparison fair, we also only compute the joint path and ignore the dynamics constraint in (\ref{eqn::linear_double_integrator_dyanmics}) for this benchmark. The initialization for the IPOPT baseline and the trajectory splitting algorithms is set to be a straight line.

Table. \ref{table1} shows the benchmark results. TrajSplit\_3 and TrajSplit\_5 denote the proposed splitting methods that split a trajectory into $3$ and $5$ segments respectively. Sampling-based planners, such as RRT, and LBKPIECE performs well in these tasks. They all achieve a $100\%$ success rate in this setting. However, since those sampling planners are stochastic, they may generate different paths for the same problem, and the computation time and solution quality also has a very large variance (for example, RRT is normally efficient to deal with those problems, but sometimes it may take over $3.7$ seconds to obtain a sub-optimal solution). We noticed in our benchmarking, that the sampling-based planners sometimes generated non-intuitive motion with much larger average path lengths than the optimization-based planners. 
RRT* is a sampling-based method that tries to deal with the optimality problem. However, for all our tasks, RRT* was not able to find an optimal solution within the $5$ seconds limit and the solution quality was qualitatively poor relative to the other optimization-based planners we compared. Similar results have been reported in other paper~\cite{meijer2017performance}. These data corroborate that RRT* is not as competitive in terms of computation efficiency.

The CHOMP algorithm in MoveIt was able to obtain optimal paths efficiently. However, the robustness of the optimization-based planner was low. In our benchmark, the planner was often caught in a local minimum and incapable of finding a feasible solution when the start pose was close to the obstacle. In contrast, the IPOPT baseline was slower but more robust, and able to solve most of the problems, which was expected as IPOPT is built for general NLP and not for speed. Our proposed trajectory splitting method demonstrated a nice balance between speed and solution quality when the splitting error tolerance was set to $10$ degrees and split into three segments. The splitting error is negligible considering the robot has 6 DOF. Trade-offs in quality and speed were straightforward by adjusting the number of splits and tolerance of the splitting error. However, the most efficient setup required balancing the number of segments that may benefit the computation time for a single primal update with additional ADMM iterations that may be required to meet the stopping criterion. The typical failure case of the proposed trajectory splitting algorithm comes from the misalignment of the splitting point that penetrates the obstacle. Due to the splitting of the trajectory, each segment tends to keep away from the obstacle. When the obstacle is thinner than the distance that results from the splitting error tolerance, the intermediate trajectory may penetrate the obstacle. We will address this problem in future work.

We further tested the behavior of our method by tuning hyperparameters: number of segments and splitting tolerance. We illustrate the variation of the computation time and path length for different splitting tolerance and different splitting segment numbers. As shown in Fig.~\ref{fig:splitting_error_length}, splitting a trajectory into more pieces resulted in a longer path length due to the splitting residual. In practice, there was a trade-off between the computation time and the splitting residual tolerance as shown in Fig.~\ref{fig:splitting_error_time}. 
To balance the solution quality and the efficiency, people may need to carefully tune these two parameters for different types of requirements.

\subsection{Experiments}
\label{subsection::Experiments}

\begin{figure*}[!ht]
\centerline{\includegraphics[scale = 0.44]{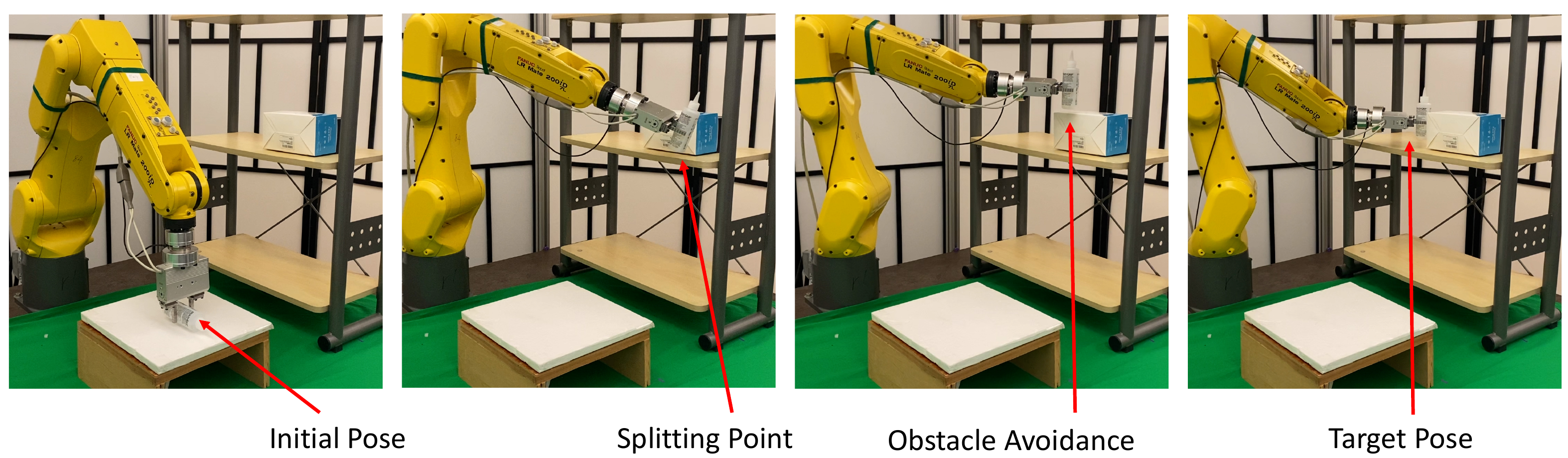}}
\caption{Experiment snapshots. The robot picks a bottle onto the bookshelf. Using the proposed trajectory splitting framework, the problem is split into two pieces and solved in a distributed manner. 
The splitting happens in the middle of the trajectory, and we can observe a tiny splitting residual during the execution in the experiment video.
}
\label{fig_experiment}
\end{figure*}

We tested the effectiveness of the trajectory splitting algorithm in a real-world scenario using a 6 DOF FANUC LRMATE 200id robot. As shown in Fig.~\ref{fig_experiment}, the robot picks a bottle from a wooden chair and places it into a bookshelf. The trajectory was initialized with linear interpolation in joint space and the problem was split into two segments with a splitting tolerance of $10$ degrees. The solution was obtained from Alg.~\ref{alg_TrajSplit}, and the assembled trajectory was sent as a list of joint space positions to the robot for execution. The snapshots from the video show the continuity of the complete trajectory, free from any collision. Note that with this splitting error tolerance a small motion artifact is observable between the two split segments.

\section{Conclusions}


Modern computational hardware is replete with opportunities for parallel computation, e.g. multi-core CPUs, GPUs, and TPUs. Our algorithm offers a novel scheme to exploit parallelism in trajectory optimization and a framework for balancing trajectory quality with computational speed. Planning tasks that require fast solve times can choose fewer segments and looser splitting tolerances to obtain quick, reasonable quality solutions. Conversely, high-accuracy trajectories can be obtained with increased segments and tighter splitting tolerance with modest increases in planning time.

For the limitations, the current implementation requires tuning of hyper-parameters. Furthermore, tight-tolerance for reducing jumps in the trajectory at the split points currently requires many more ADMM updates to achieve consensus, which can eat away gains made through parallelizing the subproblems. In future works, we plan to incorporate advances from Fast ADMM~\cite{goldstein2014fast} to address slow consensus convergence and explore using different optimizers for solving the subproblems to further improve robustness and speed.

\bibliographystyle{IEEEtran}
\bibliography{Reference}	
\vspace{12pt}
\end{document}